# Foundations of Modern Language Resource Archives

Peter Wittenburg, Daan Broeder, Wolfgang Klein, Stephen Levinson, Laurent Romary

Max-Planck-Institute for Psycholinguistics
Wundtlaan 1, 6525 XD Nijmegen
{peter.wittenburg, daan.broeder, wolfgang.klein,stephen.levinson}@mpi.nl, laurent.romary@loria.fr

**Abstract**
A number of serious reasons will convince an increasing amount of researchers to store their relevant material in centers which we will call "language resource archives". They combine the duty of taking care of long-term preservation as well as the task to give access to their material to different user groups. Access here is meant in the sense that an active interaction with the data will be made possible to support the integration of new data, new versions or commentaries of all sort. Modern Language Resource Archives will have to adhere to a number of basic principles to fulfill all requirements and they will have to be involved in federations to create joint language resource domains making it even more simple for the researchers to access the data. This paper makes an attempt to formulate the essential pillars language resource archives have to adhere to.

## 1. Introduction

The introduction of digital technology has fundamentally changed the ways in which we produce, store and use language resources. Traditionally, the focus was on creating publications as the result of the individual researcher's work and distributing them to share knowledge. These publications were enriched by examples the individual researcher found in his/her recordings and notes, the raw material not in general being accessible to the field. Now we have the situation that it is much easier to create digital audio or video recordings and to make them available at all steps of the scientific analysis process. In addition, large amounts of primary texts are available to the researchers via the Web and by harvesting newspaper and journal texts and digitized books. Indeed, the web is a gigantic source for language-oriented researchers and it will include an increasing amount of multimedia resources due to the preferences of the young generations. However, the web is focused on mainstream languages and language usages, i.e. it lacks most of the existing 6500 languages, many of which are highly endangered. It also lacks specific recordings where multimodal utterances are generated under controlled circumstances etc. Therefore, the creation of additional resources will remain crucial in the language research and documentation process.

This revolutionary change in data storage and retrieval possibilities has basically taken place in less then two decades and it is leading to a huge amount of primary data. We also see that the percentage of resources that are annotated, and thus which can be evaluated in terms of their scientific relevance, becomes increasingly smaller. A typical problem that traditional archives have re-occurs: repositories of digital data will contain an increasing amount of material that has not been enriched by 'value added' information in any substantial detail. The unprecedented growth of computer power and storage capacity creates the illusion for all participants in the data collection and analysis process that it is possible to manage an unlimited number of resources without additional efforts. In this paper we argue that this assumption will actually lead to the loss or inaccessibility of much of the data in a very short time.

A few examples may illustrate the utterly problematic situation. An investigation carried out by D. Schüller [1] in the aegis of an UNESCO project revealed that about 80% of our ethnologically motivated recordings about cultures and languages are endangered due to lack of care for the primary records by individuals or projects. We know that huge amounts of linguistically useful data is stored on private PCs encapsulated in some database with a high chance that this data will be lost when the PCs or the software will be retired or updated. Most of the websites that are used for research purposes are fragile, i.e. they will not be maintained for a longer time because funding for the project stopped or people with essential knowledge moved on. Also the lack of resource descriptions is an issue of sufficient specificity and in reusable formats is an issue. At the MPI for Psycholinguistics we had an increase of 4 TB of digital data within one year amounting to in total 15 TB. More than half of the recordings are not described by metadata, that is, there is no record of even which language is being spoken, let alone under what conditions it was recorded.

## 2. Language Resource Archives

Therefore, there has been an international trend to setup "centers" that are meant primarily to store all the data, which have a scientific or societal value even if they are no more than snapshots of the web documenting current language usage, or which have to be maintained simply for reference purposes. We will call these centers that store language resources, have expertise about their content and that give access services "language resource archives". One of their main objectives is to take care of long-term preservation of the data which makes them true archives in the traditional sense. However, as the Technical Board of IASA [2] stated correctly, it is not the task of digital archives anymore to store physical objects such as tapes and CDROMs. Since digital representations can be copied without losing information and since copying can be comparatively inexpensive the situation has changed completely: it is the content and not its specific physical existence that has to be preserved. This is in particular true where we are not forced to apply lossy compression techniques and when we take care that the digital representations are complete copies.

Due to a fundamental physical law which says that we will adversely affect objects whenever we touch them, traditional archives have to impose a very restrictive access policy. In the digital domain we argue that accessing the content does not change it, which is correct if we strictly follow the stand-off annotation rules [3] and/or apply a suitable versioning system. So digital language resource archives are expected to give easy access to the material they store. This aspect is still in serious debate, but discussions within big national libraries such as the Royal Dutch Library [4] show that even such big institutions are busy adapting their business models towards more interactive access scenarios. Of course, access here is not meant in the more traditional way that institutions such as ELDA [5] provide them at this moment. They support a web catalogue and the user can ask for the distribution of the selected resource. Again, by 'access' we do not have in mind that resource providers offer a very restricted web-based interface with the help of which one can carry out restricted queries and access singular items.

- Based on this we can summarize what can be seen as major tasks of modern language resource archives (LRA):
- LRA have to take care of long-term preservation of the hosted data and of the stability of references.
- LRA have to offer services that allow flexible access to the data according to the needs of the potential users, and permit uploading new versions and flexibly extending them.
- LRA have to offer possibilities to enrich the data, i.e. to add new resources, commentaries and relations or update existing ones. This, may of course, not influence the archived content.
- LRA have to take care that ethical and legal constraints as well as intellectual property rights aspects are taken seriously.

LRA are service centers that address the needs of the different user groups. In the first instance, the needs of researchers have to be satisfied. In some cases also the access by native speaker communities is of high relevance. But also there are students, teachers, journalists and other groups that can be mentioned as potential user groups. Satisfying all the needs would require a whole spectrum of services that a single LRA cannot meet. Therefore, an LRA has to offer appropriate open interfaces for other service providers. The services of an LRA are amongst two extremes: very shallow, in the sense that they e.g. expose the metadata or content to simple search engines or, more deeply, rich data that offers interfaces for programmers.

## 3. Principles of Language Resource Archives

Based on what we have described so far, we can describe a number of principles that have to be met by modern digital Language Resource Archives. These principles, have as a corollary, that they imply requirements for technologies to be applied to them.

1. LRA have to implement a strategy for long-term preservation that includes a migration plan to new technologies and a distribution plan to create copies of the data at different locations following different protocols. This requires a kind of low-level federation, since you can only exchange sensitive data with trusted servers and organizations. This federation implies both agreements on the technology level (exchange protocols etc) as agreements in the ethical and judicial domain.

2. LRA have to adopt as much as possible widely used and open standards for all data including the metadata and relations between the resources. A conversion will be necessary towards these standards which includes structure descriptions for textual data that are compliant with generic schemas. Finally, the degree of coherence and compliance to such schemas will influence the costs of migration towards new representation formats that will emerge.

3. LRA have to differentiate between physical storage structure, which is characterized by servers, disks etc., and the linguistic archive organization, which is characterized by resource metadata and linguistically meaningful categorizations. While the first is defined by system managers and influenced by technological considerations and therefore changing frequently, the latter is determined by scientific considerations and comparatively stable. Archive management, resource discovery and usage should make use of the linguistic organization.

4. LRA have to agree on mechanisms that are able to resolve Unique Resource Identifiers (URIDs) to physical paths. Only the use of URIDs will allow us to maintain stable references and to make a distinction between an archival object and its many instances (copies) that can exist at other archives. Stable references to digital resources will become increasingly important since publications will increasingly often refer to them and are indispensable now when we want to create an interlinked domain of language resources.

5. LRA have to devise a strategy to allow selected users to upload new resources to an archive or to update existing resources without destroying the existing ones. This will require a web-based upload and management system offering work spaces and a smart versioning mechanism. It is one of the basic principles of archiving that archived data may not be touched. In the digital era this could be disastrous, since there may be references to old resources and these have to be resolved to the original objects even when new versions are available.

6. LRA must offer a powerful access management system that allows us to define access policies and offers delegation mechanisms. This is important to give depositors full control of granting access to "their" data. Relevant material will only be deposited, if the archivist declares to respect the rights of the creators and guarantees that they know that they always can access their material.

7. LRA must offer different layers of access to the data dependent on the expected user groups. This is a very problematic point since we often cannot anticipate what kind of user interface special user groups such as for example members of language communities expect. The access techniques range from geographical browsing, metadata browsing and searching to more advanced methods to access complex linguistic types such as annotated media files and multimedia lexica. Most important for an archive is to offer neutral access mechanisms that allow the user to access the individual resource without any embedding if this is required. For language technology users and to allow setting up local data centers it is often required to also offer the download of a complete sub-archive including all bundling and metadata information.

8. LRA will have to offer ontology support in the future to compensate for the linguistic encoding differences. LRA house contributions from various individuals and projects all using different terminologies to describe linguistic phenomena. Users will want to carry out for example searches across different corpora which will only work when there is smart ontology support.

9. In the future LRA will also have to offer services that allow selected users to add comments to fragments and to mark relations between them. These enrichments are part of the archive, i.e., they have to be stored in open formats including the bundling information as well. However, the original resource may not be affected.

Most – if not all - of the current repositories housing language resource data do not operate according to these principles yet. However, the pressure to do so will increase. LRA, if they are to survive in a competitive domain, will have to operate at a cost-effective level and nevertheless offer smart and stable services to the different user groups. LRA can be part of different scenarios to guarantee persistence: they can offer all services themselves, i.e., take care of redundant storage and appropriate migration strategies and access services on the one extreme end or use computer centers of libraries to take care of long-term storage and limit their own activities to providing access services. Yet we cannot rely on the services of traditional libraries and archives since they lack the knowledge about the content and have no experience with modern access scenarios as described in this paper.

## 4. Archive Federations

LRA will have to become members of archive federations, i.e., communities of trust and virtual integration. The term "federation" covers technological and in particular organizational and juridical aspects. In the domain of language resources we can see two related initiatives to create a federation of archives. DELAMAN [6] is an international network of archives housing endangered language and music material. Two of the major goals of this network are (1) to create a community of mutual trust based on an agreed ethical and juridical framework that will allow us to exchange data and (2) to understand the technologies that allow us to create a joint access domain. The reason for focusing on these two aspects are the necessity of improving the conditions for the long-term preservation of the stored unique material and the knowledge that different archives host material about the same languages or those that are spoken in neighboring communities. Researchers want to see all resources of a specific language or want to study the influences between languages without being bothered by all kinds of organizational and technical boundaries.

DAM-LR (Distributed Access Management for Language Resources) [7] is a European project where four archives serving different communities such as fieldworkers, phoneticians and computational linguists are taking practical steps to come to a joint virtual archive. All four partners have been investing substantial funds to form full-fledged language resource archives according to the above mentioned principles. The project has already worked out solutions for the essential pillars of an archive federation and is currently busy implementing them:

(1) establishing a domain of trusted servers and services by setting up a PKI system [8] based on EUGridPMA certificates [9] (this mechanism is supported world-wide);
(2) establishing a joint domain of metadata by making use of the IMDI metadata infrastructure [10] (due to the support for research and management shallow metadata sets as Dublin Core are not sufficient);
(3) establishing a joint domain of Unique Resource Identifiers based on the widely used Handle System [11] where each archive manages its own URID sub-domain and therefore is free to specify the syntax of its URIDs;
(4) establishing a distributed authentication and authorization system where the authentication is left to the home institutions of users and where Shibboleth [12] is used to exchange user credentials to allow authorization.

The federation includes a number of agreements between the partners such as
a) agreeing on the user attributes that are exchanged when determining access rights;
b) associating the access rights information with URIDs and thereby assuring that the owning institute defines the access rights for all copies;
c) creating mirror sites for resolving handles;
d) using Shibboleth for exchanging about users, but leaving the decision about the authentication system itself to the partners.

The partners identified the need to develop a resource managing component that interfaces with the other components, implements the access policies defined and an advanced access specification management component that can be used by archive managers and depositors to specify policies and access permissions. All specifications for the agreements have been made and have now to be tested in reality.

Where possible, DAM-LR is relying on components that have already shown their robustness and reliability. Shibboleth will be used although we foresee that the typical scenario where authorization is done based on user classifications such as "being a member of a student class" or "belonging to a certain staff category" will not apply to most cases in our domain. A problem may emerge at large universities where the user attributes are defined at a high university level. Departments participating in DAM-LR will not be able to convince the university boards to change the rules and also store attributes specific for the DAM-LR scenario. A simple solution can be realized when for instance LDAP is applied for authentication. A local copy with filtered information could be created and the necessary attributes could be added under local responsibility.

## 5. Community State

It is obvious that modern language resource archives can only tackle the above mentioned problems, since different initiatives have driven the language resource community during the last decades. Language resource specific standardization efforts have been taken by initiatives such as TEI [13], EAGLES [14] and ISLE [15]. However, only initiatives such as ISO TC37/SC4 [16] have recognized the necessity to specify generic models and schemas. It is obvious that proposals such as LMF (Lexical Markup Framework) are needed to achieve some of the goals. We also can build upon the standards developed by Unicode [17], W3C [18], ISO [19] and OAI [20] with respect to unified character encoding, the XML language to describe document structures, unified language codes, metadata harvesting protocol and many others. With respect to building federations we can build upon the knowledge and tools developed within the digital library community and Grid initiatives such as GGF [21].

## 6. Summary and Conclusions

The language resource domain is confronted by an enormous increase of interrelated resources that have to be managed. We foresee that this task in all its respects can only be carried out by new types of centers which we call "language resource archives". These archives are dealing with digital material, where the rule that physical archives may not be touched (in order to preserve them for future generations) is not applicable anymore. Therefore, these new type of archives should offer services for extensions and enrichments, guaranteeing however, that the original content will not be affected.

In this paper, we have described a number of principles that these archives need to follow to offer smart and stable access services, including the primary task of long-term preservation. Currently, as far as we know no language resource archive fully adheres to these principles, but due to the previous standardization work and the experience in this field we are optimistic that we now can build such archives. In addition, these archives need, in the next few years, to form or affiliate to federations of archives. Currently, we know of two closely collaborating initiatives which are discussing and testing federation methods. Within a few years we expect to see the first full-fledged digital language resource archives to be operating within such federations. These will offer the researchers a much more integrated and accessible domain of language resources.